\documentclass{article}
\usepackage{ijcai21}

\usepackage{times}
\usepackage{soul}
\usepackage{url}
\usepackage[hidelinks]{hyperref}
\usepackage[utf8]{inputenc}
\usepackage[small]{caption}
\usepackage{graphicx}
\usepackage{subfig}
\usepackage{amsmath}
\usepackage{amsthm}
\usepackage{booktabs}
\usepackage{algorithm}
\usepackage{algorithmic}
\usepackage{mathrsfs}
\usepackage{amssymb}
\usepackage{color}

\newtheorem{theorem}{Theorem}
\newtheorem{lemma}{Lemma}
\title{
Hindsight Value Function for Variance Reduction in Stochastic Dynamic Environment}

\author{
Jiaming Guo$^{1,2,3}$
\and
Rui Zhang$^{1,2}$\and
Xishan Zhang$^{1,2}$\and
Shaohui Peng$^{1,2,3}$ \and
Qi Yi$^{1,2,4}$ \and
Zidong Du$^{1,2}$ \and
Xing Hu$^{1}$ \and
Qi Guo$^1$ \and
Yunji Chen$^{1,3,5*}$
\affiliations
$^1$SKL of Computer Architecture, Institute of Computing Technology, CAS, Beijing, China\\
$^2$Cambricon Technologies \\
$^3$University of Chinese Academy of Sciences, China \\
$^4$University of Science and Technology of China \\
$^5$CAS Center for Excellence in Brain Science and Intelligence Technology, CEBSIT
\emails
\{guojiaming18s, zhangrui, zhangxishan, pengshaohui18z\}@ict.ac.cn,
yiqi@mail.ustc.edu.cn,
\{duzidong, huxing, guoqi, cyj\}@ict.ac.cn
}
\begin{document}

\maketitle

\begin{abstract}
 Policy gradient methods are appealing in deep reinforcement learning but suffer from high variance of gradient estimate. To reduce the variance, the state value function is applied commonly. However, the effect of the state value function becomes limited in stochastic dynamic environments, where the unexpected state dynamics and rewards will increase the variance. In this paper, we propose to replace the state value function with a novel hindsight value function, which leverages the information from the future to reduce the variance of the gradient estimate for stochastic dynamic environments.
 %In this paper, we propose a novel hindsight value function that leverages the information from the future to replace the state value function as a baseline and reduce the variance in the stochastic dynamic environment.
 Particularly, to obtain an ideally unbiased gradient estimate, we propose an information-theoretic approach, which optimizes the embeddings of the future to be independent of previous actions. In our experiments, we apply the proposed hindsight value function in stochastic dynamic environments, including discrete-action environments and continuous-action environments. Compared with the standard state value function, the proposed hindsight value function consistently reduces the variance, stabilizes the training, and improves the eventual policy.
\end{abstract}

\section{Introduction}
%\footnote{$^{*}$Corresponding Author}
\let\thefootnote\relax\footnotetext{$^{*}$Corresponding Author}

Deep reinforcement learning has achieved impressive results on solving sequential decision-making tasks, such as video games\cite{DBLP:journals/nature/MnihKSRVBGRFOPB15}, board games\cite{DBLP:journals/nature/SilverHMGSDSAPL16}, and robotics\cite{DBLP:journals/jmlr/LevineFDA16}. 
%The policy gradient methods, a main class of algorithms for reinforcement learning, have been widely adopted because they directly optimize the policy based on cumulative reward and can straightforwardly be used with neural networks.  
%这里用不用提一下是通过gradient estimate进行policy optimize的
The policy gradient methods\cite{DBLP:journals/ml/Williams92,DBLP:conf/nips/SuttonMSM99,DBLP:conf/nips/Kakade01,DBLP:journals/corr/SchulmanWDRK17} have been widely adopted for reinforcement learning because they can be used with deep neural networks straightforwardly and directly adjust the log probability of actions by estimating gradient based on the future cumulative rewards influenced by these actions.
%(or directly optimize the policy by estmating gradient based on cumulative reward 不知道哪个好，没注释的写得清楚但是句子有点长，这个关系没有第一个直接但是比较短).
However, the gradient estimate in policy gradient methods suffers from high variance because the effect of an action on the cumulative reward is confounded with the influence of future actions, current state, and stochastic dynamics of the environment. 
One effective method to reduce variance is substracting a ``baseline'' from the cumulative reward to exclude the influence of confounding factors. 
%The most common choice of a baseline is the state value function to removes the influence of the current state.
The most common baseline is the state value function, which predicts the average performance starting from the current state\cite{DBLP:journals/tnn/SuttonB98}. 
Thus, the state value function is able to reduce the high variance of gradient estimate by removing the influence of the current state.

%However, using the state value function which only contains information about the current state as a baseline fails to exclude the influence of unexpected state dynamics and reward in a stochastic dynamic environment.
%stochastic dynamic environment which consists of unexpected state dynamics and reward. 
%Thus, when the environment becomes drastically stochastic, the effect of the state value function baseline becomes limited. 

%Let us consider an archery player learning to shoot at the target in the wind. %The player pulls the bow, aims, and releases the arrow to the target. 
%The wind will change the trajectory of the arrow and affect the result.
%Even the player gets told the relative position of target, the direction and speed of wind before the arrow is released which can be regarded as ``the current state'', improving the skill from the result is difficult as the changes of wind will also affect the result when the arrow flying.
However, the state value function will not work effectively in stochastic dynamic environments. The reason is that the state value function baseline only considers the current state and fails to exclude the influence from unexpected state dynamics and rewards in stochastic dynamic environments.
For example, when an archery player learns to shoot at the target in the ever-changing wind, the wind will change the trajectory of the arrow and affect the result.
%Even the player gets told the relative position of target, the initial direction and speed of wind which can be regarded as ``the current state'', improving the skill from the result is difficult as the unexpected changes of wind will also affect the result when the arrow flying.
Even the player gets told the relative position of the target, the initial direction and speed of the wind (which can be regarded as ``the current state''), improving the skill from the result is difficult as the unexpected changes of wind (which can be regarded as ``the stochastic dynamics of the environment'') will also affect the result when the arrow is flying.
When humans introspect to improve a certain skill, they will take the whole experience including the current state and dynamics of the environment in the future into consideration.
%Inspired by this, we draw the conclusion that the future information should be considered in reinforcement learning.
Inspired by this, we envision that the environmental dynamics should be considered for variance reduction in addition to the current state, especially when the environment becomes drastically stochastic.

%In this paper, we propose to learn a new value function in hindsight that receives future states and rewards as an additional input.
In this paper, we propose a hindsight value function that can work as a new baseline to reduce variance by removing the influence of unexpected state dynamics and rewards on the cumulative rewards. The hindsight value function receives future states and rewards as additional input and is learnt in hindsight, which is similar to the process of human introspection.
With the new hindsight value function as a baseline, the policy can be adapted efficiently to optima consequently.
%Then we can employ this hindsight value function as a new baseline to assess the quality of actions and calculate the policy gradient, the process of which is similar to human introspection. 
%This hindsight value function can be employed as a new baseline to assess the quality of actions and calculate the policy gradient, which is similar to the process of human introspection.
%As this new baseline excludes the influence of current state and unexpected state dynamics and reward, the variance of gradient estimate is further reduced. 
%This new baseline further helps variance reduction by removing the influence of unexpected state dynamics and reward on the cumulative rewards.
%This new baseline further helps credit assignment by removing the influence of unexpected state dynamics and reward on the cumulative rewards, thereby reducing variance. 
%Consequently, the policy can be adapted efficiently to optima with the hindsight value function as a baseline.
%In other words, the information about future states can provide a better evaluation of how well the agent performs.

One major challenge of designing the hindsight value function is to keep the gradient estimate unbiased. Directly adopting the future states and rewards to compute the baseline will yield a biased gradient estimate, due to the coupling relationship between future states\&rewards and the current action. 
Therefore, we propose a methodology to decouple current actions from the future information based on the guidance from information theory.
%The challenge of designing the hindsight value function is to keep the gradient estimate bias-free. 
%Note that the future states and rewards contain the information of the current action. 
%Thus, they can not be directly used to compute the baseline as when computing the policy gradient it will yield a biased gradient estimator.
%Note that the future states and rewards contain the information of the current action, they will yield a biased gradient estimator if they are directly used to compute the baseline. 
%As humans can distinguish whether something happens because of random factors or something they did before, we filter out the information about current actions from the future. %based on the guidance from information theory. 
The information from the future is provided by the future embeddings, which are supposed to be independent of the current actions. 
To accomplish the independence, we minimize the upper bound of mutual information between future embeddings and the current action towards zero. 
The hindsight value function takes embeddings of the future as input and leads to an ideally unbiased gradient estimator.

 Experimentally, we compare our hindsight value function to the standard state value function by applying them to the policy gradient methods A2C\cite{DBLP:conf/icml/MnihBMGLHSK16} and PPO\cite{DBLP:journals/corr/SchulmanWDRK17} for environments including \textit{gridworld} and the robotics control benchmark in the MuJoCo physics simulator\cite{DBLP:conf/iros/TodorovET12}. Our results show that the proposed hindsight value function consistently reduces the variance, stabilizes the training, and improves the eventual policies.

\section{Related Works}
%In reinforcement learning, two main classes of algorithms are value-based methods and policy-based methods. Value-based methods maintain an estimate of the value of performing each action in each state, and choose the actions associated with the most value in their current state. In contrast, policy-based methods maintain an explicit policy and directly draw actions from the policy.

\paragraph{Policy gradient.} Policy gradient methods\cite{DBLP:conf/nips/SuttonMSM99} are a subset of policy-based methods that maintain an explicit policy and directly draw actions from the policy. These methods represent the policy using a differentiable parameterized function approximator and use stochastic gradient ascent to update its parameters to achieve more reward. Many researchers have proposed a series of variety of policy gradient methods including Advantage Actor-Critic(A2C), Asynchronous Advantage Actor-Critic(A3C)\cite{DBLP:conf/icml/MnihBMGLHSK16},Trust Region Policy Optimization(TRPO)\cite{DBLP:conf/icml/SchulmanLAJM15}, Proximal Policy Optimization(PPO)\cite{DBLP:journals/corr/SchulmanWDRK17} and so on. We mainly carry out the experiment based on the commonly used method A2C and PPO.

\paragraph{Variance reduction.} The major drawback of policy gradient methods is the high variance in gradient estimate, and a large body of work has investigated variance reduction techniques. Reduce variance through using a baseline has been shown to be effective. %These works use the state value function as baseline which only uses the information of the current state while we leverage the information of future. 
\cite{DBLP:conf/iclr/WuRDKBKMA18} 
consider MDPs with multi-variate independent actions and propose the state-action-dependent baseline to improve training efficiency by explicitly factoring out the effect of other actions for each action.
\cite{DBLP:conf/iclr/MaoVSA19} introduces a kind of environments in which state dynamics depend on the input process. They propose an input-dependent baseline to reduce the variance and use a meta-learning approach to learn the baseline. Q-Prop\cite{DBLP:conf/iclr/GuLGTL17} makes use of an action-dependent control variate to reduce the variance. Some works consider the bias-variance tradeoff in policy gradient methods. GAE\cite{DBLP:journals/corr/SchulmanMLJA15} replaces the Monte Carlo return with a $\lambda$-weighted return estimation with value function bootstrap. However, when the state value function provides limited information, the bootstrap will not work. We aim to reduce the variance by providing a stable and informative value function as baseline.% We provide a stable and informative hindsight value function that can be combined with bootstrapping.
 
\paragraph{Mutual information.} Mutual information (MI) is a fundamental measurement of the dependence between two random variables. It has been applied to a wide range
of tasks in machine learning, including generative modeling\cite{DBLP:conf/nips/ChenCDHSSA16}, information bottleneck\cite{DBLP:journals/corr/physics-0004057}, and domain adaptation\cite{DBLP:journals/tip/GholamiSRBP20}. Some works applied the MI in reinforcement learning. \cite{DBLP:conf/nips/TaoFP20} use the mutual information to learn the representation of states to facilitate efficient exploration. DADS\cite{DBLP:conf/iclr/SharmaGLKH20} use mutual information to discover skills that are identifiable by the transitions of state.
In our proposed method, we utilize MI to measure the dependence between future embedding and previous actions. As far as we know, we are the first to use MI to learn a value function.

\section{Preliminaries}
\subsection{Notation}
This paper assumes an agent interacts with the environment over discrete timesteps, modeled as a Markov Decision Process(MDP), defined by the 6-tuple ($\mathcal{S},\mathcal{A},\mathcal{P},\rho_0,r,\gamma$). In this setting, $\mathcal{S} \subseteq \mathbb{R}^n$ is a set of n-dimensional states, $\mathcal{A} \subseteq \mathbb{R}^m$ is a set of m-dimensional actions, $\mathcal{P}:\mathcal{S} \times \mathcal{A} \times \mathcal{S} \rightarrow [0,1]$ is the state transition probability distribution, $\rho_0: \mathcal{S} \rightarrow [0,1]$ is the distribution over initial states, $r: \mathcal{S} \times \mathcal{A} \rightarrow \mathbb{R}$ is the reward function, and $\gamma \in (0,1)$ is the per timestep discount factor. We denote a stochastic policy as $\pi_\theta: \mathcal{S} \times \mathcal{A} \rightarrow [0,1]$ which is parameterized by $\theta$. We aim to optimize the expected return $\eta(\pi)= \mathbb{E}_\tau[\sum^\infty_{t=0}\gamma^t r(s_t,a_t)]$, where $\tau = (s_0,a_0,\dots)$ is the trajectory following $s_0 \sim \rho_0, a_t \sim \pi(a_t|s_t), s_{t+1} \sim \mathcal{P}(s_{t+1}|s_t,a_t)$. We define the value function as $V(s_t) = \mathbb{E}_{a_t,s_{t+1},\dots}[\sum_{l=0}^\infty \gamma^l r(s_{t+l},a_{t+l})|s_t]$.
\subsection{Policy Gradient}
The Policy Gradient Theorem\cite{DBLP:conf/nips/SuttonMSM99} states that 
$$
    \nabla_\theta\eta(\pi_\theta) = \mathbb{E}_{\tau \sim \pi_\theta}[\sum_{t=0}^{\infty}\nabla_\theta log \pi_\theta(a_t|s_t) \sum_{t^{'}=t}^\infty \gamma^{t^{'}-t}r_{t^{'}}]  
$$
For convenience, we define $R_{t}(\tau)=\sum_{t^{'}=t}^\infty \gamma^{t^{'}-t}r_{t^{'}}$ and $\rho_\pi (s) = \sum_{t=0}^{\infty}\gamma^t p(s_t=s)$ as the state visitation frequency. As using Monte Carlo estimation for  $R_{t}(\tau)$ suffers from high variance, we can subtract off a quantity dependent on $s_t$ from it to reduce the variance of this gradient estimator without introducing bias. The policy gradient estimation with a baseline becomes
$\nabla_\theta\eta(\pi_\theta) = \mathbb{E}_{\tau \sim \pi_\theta,\rho_\pi}[\nabla_\theta log \pi_\theta(a|s) (R_{t}(\tau)-b(s_t))]$. The baseline is usually the state value function $V(s_t)$. In practice, a parametric function $v_\theta$ is often optimized to estimate the value. However, the optimization  may be difficult in domains with high-dimensional
observation spaces. Things become worse when environments or rewards are strongly stochastic.
    
%\subsection{Value Estimation}
%Value estimation is a critical component of reinforcement learning. For policy gradient methods, if a baseline is employed to reduce the variance of gradient estimation, the optimal baseline is derived to be\cite{DBLP:conf/nips/GreensmithBB01}:
%$$
%    b^* = \frac{\mathbb{E}_{\tau \sim \pi_\theta,\rho_\pi}[\nabla_\theta log \pi_\theta(a_t|s_t)^T\nabla_\theta log \pi_\theta(a_t|s_t)R(\tau)]}{\mathbb{E}_{\tau \sim \pi_\theta,\rho_\pi}[\nabla_\theta log \pi_\theta(a_t|s_t)^T\nabla_\theta log \pi_\theta(a_t|s_t)]}
%$$
%As this baseline is hard to estimate, it is usually replaced by $\mathbb{E}[R(\tau)]$ which can be represented as $V(s_t)$ as above. An accurate value estimate is critical as it means lower variance in the gradient estimate.

%In practice, a parametric function $v_\theta$ is often employed to estimate the value. Trajectories and returns are sampled under the current policy. Then the parameters of $v_\theta$ are updated based on these samples. This regression may be difficult in domains with high-dimensional
%observation spaces. Things become worse when environments or rewards
%are strongly stochastic.

\section{Hindsight Value Function}
\label{hindsight}
  We introduce a hindsight value function $v_h$, which represents the expected return conditioned on not only current state $s_t$ but also future states and rewards. The hindsight value function can be represented as 
$$
  v^h(s_t,s_+,r_+)=\mathbb{E}[R|s_t,r_t,s_{t+1},r_{t+1}\dots]
$$
  where $s_+$ means the future states and $r_+$ the future rewards. However, the expected return seems a constant $R(\tau)$ when the whole future is known. Then the policy gradient becomes biased as $ \nabla_\theta\eta(\pi_\theta)=\mathbb{E}_{\tau \sim \pi_\theta,\rho_\pi}[\nabla_\theta log \pi_\theta(a|s) (R_{t}(\tau)-R_{t}(\tau))]$ = 0.
\subsection{Embeddings of Future}
  To tackle the problem mentioned above, we have to modify the hindsight value function $v^h$. It should exploit the information from the future but keep the policy gradient unbiased when employed as a baseline. We introduce a set of hindsight vectors $h^+:(h_{t+1},h_{t+2},\dots)$, the embeddings of the future, which distill information from the future states and rewards. The hindsight value function takes these vectors as input rather than directly exploiting the future state-reward pairs. Then the hindsight value  function can be written as:
    \begin{equation}
  v^h(s_t,h^{+})=\mathbb{E}[R|s_t,h_{t+1},h_{t+2}\dots].
  \end{equation}
\begin{figure*}[t]
  \centering
  \includegraphics[width=0.83\linewidth]{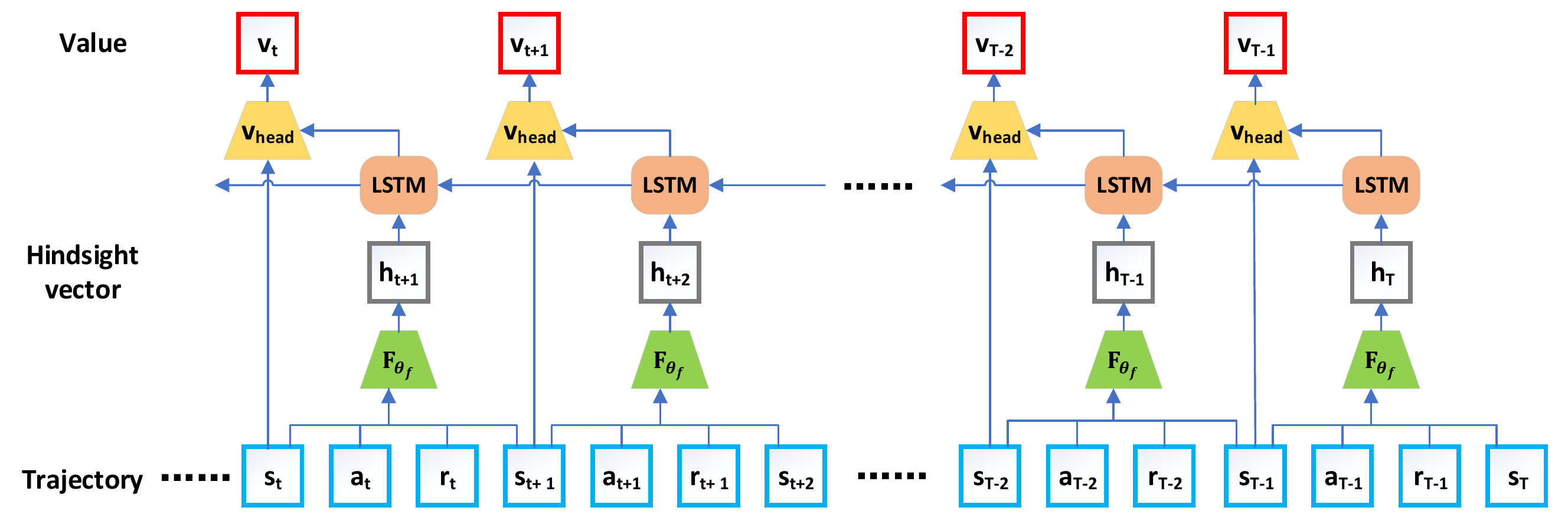}
 
  \caption{Model architecture of hindsight value function.}
  \label{lstm}
 
\end{figure*}  
We regard $h_t$ as a hindsight vector that contains the information of a single time step of the future. The hindsight vector should satisfy two properties.  First, the hindsight vector should be irrelevant to the actions of the agent to guarantee the gradient estimate unbiased. Second, the hindsight vector is expected to contain enough information of the future for $v^h$ to estimate the expected return.
Intuitively, the hindsight vector can be explained as the changing wind in archery, the future traffic when driving, or abstractly, just random factors in some cases.

The hindsight vector $h$ is designed to be a function of the current state-action pair $(s,a)$ and the future state-reward pair $(s',r)$: $h=F((s,a),(s',r))$. The hindsight vector $h$ should satisfy the constraint:
\begin{equation}
  I(h;(s,a)) = 0, \label{dependence}
\end{equation}
where $I[\cdot;\cdot]$ is the \textit{Mutual information}(MI) between two random variables. Note that Equation \ref{dependence} is satisfied if and only if $h$ and $(s,a)$ are independent. We analyze the hindsight vector and claim that under this constraint, the gradient estimate is unbiased with $v^h(s_t,h^+)$ as a baseline. First, we state a lemma to show that under the MDP definition, the hindsight vector $h_{t+1:\infty}$ is independent of previous actions $a_t$. 

\begin{lemma}
\label{independent}
For a Markov Decision Process, ${\forall}$  k $\in$ $\mathcal{N}_+$, $I(h_{t+k},a_t) = 0$, i.e., $h_{t+k}$ and $a_t$ are independent.

\proof

For k = 1, $h_{t+1}$ and $a_t$ are independent according to Equation \ref{dependence}.

For $k > 1$, we have two equivalent forms of expansion for the mutual information:
$I(h_{t+k};(s_{t+k-1},a_{t+k-1}),a_t) \\= I(h_{t+k};(s_{t+k-1},a_{t+k-1}) +
I(h_{t+k};a_t|(s_{t+k-1},a_{t+k-1}))\\
=I(h_{t+k};a_t)+I(h_{t+k};(s_{t+k-1},a_{t+k-1})|a_t)$.

As $h_{t+k}$ is a deterministic function of state-action pair $(s_{t+k-1},a_{t+k-1})$ and next state-reward pair $(s_{t+k},r_{t+k-1})$, according to the property of MDP, $I(h_{t+k};a_t|(s_{t+k-1},a_{t+k-1})) = 0$. So we have,

$I(h_{t+k};(s_{t+k-1},a_{t+k-1})\\
=I(h_{t+k};a_t)+I(h_{t+k};(s_{t+k-1},a_{t+k-1})|a_t)$\\
$\Rightarrow I(h_{t+k};a_t) \leq I(h_{t+k};(s_{t+k-1},a_{t+k-1})=0$.

Then we get the conclusion that $h_{t+k}$ and $a_t$ are independent.
\end{lemma}
Using Lemma \ref{independent}, we prove that the gradient estimate keeps unbiased when the hindsight baseline takes hindsight vectors $h^+$ as input.

\begin{theorem}
The hindsight baseline does not bias the gradient estimate.

\proof
$\nabla_\theta\eta(\pi_\theta) \\
=\mathbb{E}_{\tau \sim \pi_\theta,\rho_\pi}[\nabla_\theta log \pi_\theta(a|s) (R_{t}(\tau)-v(s_t,h^+))]\\
= \mathbb{E}_{\tau \sim \pi_\theta,\rho_\pi}[\nabla_\theta log \pi_\theta(a|s) (R_{t}(\tau))] \\
- \mathbb{E}_{\tau \sim \pi_\theta,\rho_\pi}[\nabla_\theta log \pi_\theta(a|s) v(s_t,h^+)]$.

Therefore, we only need to show\\
$\mathbb{E}_{\tau \sim \pi_\theta,\rho_\pi}[\nabla_\theta log \pi_\theta(a|s) v(s_t,\textbf{h})]=0$:

$\mathbb{E}_{\tau \sim \pi_\theta,\rho_\pi}[\nabla_\theta log \pi_\theta(a|s) v(s_t,h^+)]\\
=\mathbb{E}_{s_{0:t},a_{0:t-1}}[\mathbb{E}_{s_{t+1:T},a_{t:T-1}}[\nabla_\theta log \pi_\theta(a|s) v(s_t,h^+)]]\\
=\mathbb{E}_{s_{0:t},a_{0:t-1}}[\mathbb{E}_{s_{t+1:T},a_{t:T-1}}[v(s_t,h^+)]\\\mathbb{E}_{s_{t+1:T},a_{t:T-1}}[\nabla_\theta log \pi_\theta(a|s)]]\\
=\mathbb{E}_{s_{0:t},a_{0:t-1}}[\mathbb{E}_{s_{t+1:T},a_{t:T-1}}[v(s_t,h^+)]\mathbb{E}_{a_t}[\nabla_\theta log \pi_\theta(a|s)]\\
=\mathbb{E}_{s_{0:t},a_{0:t-1}}[\mathbb{E}_{s_{t+1:T},a_{t:T-1}}[v(s_t,h^+)] \cdot 0]=0
$.
\end{theorem}

 The hindsight vectors are also supposed to contain enough information for $v_h$ to approximate the expected reward. We achieve this by maximizing the mutual information between $h$ and $(s',r)$ conditioned on $(s,a)$:
 \begin{equation}
     max(I(h;(s',r)|(s,a))).
 \end{equation}

In practice, we employ a neural network $F_{\theta_f}$ parameterized by $\theta_f$  to obtain the hindsight vectors: $h=F_{\theta_f}((s,a),(s',r))$. However, we can not make sure the constraints talked above holds when $F_{\theta_f}$ is a neural network. We transfer these constraints to the following functional:
 \begin{equation}
 \label{functional}
   \mathcal{L}(\theta_f) = I(h;(s,a)) - I(h;(s',r)|(s,a)).
 \end{equation}

\subsection{Learning the Hindsight Vector}

We illustrate the method to minimize the target functional given by Equation \ref{functional}. We first aim to maximize the mutual information $I(h;(s',r)|(s,a))$  that is related to the information contained in $h$. Using the entropy form, this term can be represented as:
$$
H((s',r)|(s,a))-H((s',r)|(s,a),h).
$$
Thus, we only need to minimize the conditional entropy $H((s',r)|(s,a),h)$. This can be transformed to a prediction task.
We employed a neural network $P_{\theta_p}$ which takes $h$ and $(s,a)$ as input and predicts the state-reward pair $(s',r)$. The loss function is following:
\begin{equation}
\label{loss1}
    L_P(\theta_f,\theta_p)=\|P_{\theta_p}(h,(s,a))-(s',r)\|_2^2.
\end{equation}

Now let us consider the first term of Equation \ref{functional}. To minimize the mutual information, we employ the \textit{contrastive log-ratio upper bound}(CLUB)\cite{DBLP:conf/icml/ChengHDLGC20}, which is a method to estimate the upper bound of mutual information. The CLUB method estimates the mutual information by the difference of conditional probabilities
between positive and negative sample pairs. For random variables \textbf{x} and \textbf{y}, with the conditional distribution $p(\textbf{y}|\textbf{x})$, the mutual information contrastive log-ratio upper bound is defined as:
$$
I_{CLUB}(\textbf{x};\textbf{y}) = \mathbb{E}_{p(\textbf{y},\textbf{x})}[log p(\textbf{y}|\textbf{x})]-\mathbb{E}_{p(\textbf{x})}\mathbb{E}_{p(\textbf{y})}[log p(\textbf{y}|\textbf{x})].
$$
However, in our case, the conditional probability $p(h|\\
(s,a))$ is unknown. Thus, a variational distribution $C_{\theta_c}(h|
(s,a))$ is introduced to approximate the conditional probability and replace 
$p(h|(s,a))$ in $I_{CLUB}$.  Consequently, a variational CLUB term of $h$ and $(s,a)$ is defined as:
\begin{equation}
\begin{aligned}
  I_{vCLUB}((s,a);h)
  = \mathbb{E}_{h,(s,a))}[log C_{\theta_c}(h|(s,a))]\\
  -\mathbb{E}_{p((s,a))}\mathbb{E}_{p(h)}[log C_{\theta_c}(h|(s,a))].  
  \end{aligned}
\end{equation}

The unbiased estimate for $I_{vCLUB}((s,a);h)$ is obtained by sampling. Then we reduce the mutual information term by optimizing the loss function:
\begin{equation}
\label{loss2}
     L_F(\theta_f)= I_{vCLUB}((s,a);h).
\end{equation}

We now illustrate how to optimize the hindsight vector encoding function $F_{\theta_f}$ according to loss functions given in Equation \ref{loss1} and Equation \ref{loss2}. When the agent explores in the environment, we store the tuple $(s,a,s',r)$ into a buffer $\mathcal{B}$. At each training iteration, we first sample a batch of tuples $\{(s_i,a_i,s'_i,r_i)\}$ from the buffer and calculate the corresponding $h_i$ by $F_{\theta_f}$. Then we update the variational distribution $C_{\theta_c}(h|(s,a))$, by maximizing the log-likelihood $L(\theta_c)=\frac{1}{N}\sum_{i=1}^N log C_{\theta_c}(h_i|(s_i,a_i))$. After $C_{\theta_c}(h|(s,a))$ is updated, we calculate the $L_F(\theta_f)$ and $L_P(\theta_f,\theta_p)$. Finally, the gradient is calculated and back-propagated to $\theta_f$ and $\theta_p$. During training, the hindsight vector $h$ will contain more information to facilitate the value estimate and the hindsight value function will update towards bias-free.

\begin{algorithm}[tb]
\caption{Learning hindsight vector}
\label{alg1}
\textbf{Input}: Buffer $\mathcal{B}$ containing tuples $(s_t,a_t,s',r_t)$, predict net $P_{\theta_p}$ with paremeter $\theta_p$, variational distribution network $C_{\theta_c}$ with parameter $\theta_c$, hindsight vector encoding network $F_{\theta_f}$ with paremeter $\theta_f$, batch size N.
\begin{algorithmic}[1] %[1] enables line numbers
\STATE Sample $\{(s_i,a_i,s'_i,r_i)\}_{i=1}^N$ from $\mathcal{B}$
\FOR{$i=1$ to $N$}
\STATE Calculate hindsight vector $h_i=F_{\theta_f}(s_i,a_i,s'_i,r_i)$
\ENDFOR
\STATE Log-likelihood $\mathcal{L}(\theta_c)= \frac{1}{N} \sum_{i=1}^N log C_{\theta_c}(h_i|(s_i,a_i))$
\STATE Update $C_{\theta_c}$ by maximizing $\mathcal{L}(\theta_c)$
\FOR{$i=1$ to $N$}
\STATE Sample $k'_i$ uniformly from $\{1,2,\dots,N\}$
\STATE $U_i=log C_{\theta_c}(h_i|(s_i,a_i))-log C_{\theta_c}(h_{k'_i}|(s_i,a_i))$
\ENDFOR
\STATE Variantional contrastive log-ratio upper bound \\$I_{vCLUB} = \frac{1}{N}\sum_{i=1}^N U_i$
\FOR{$i=1$ to $N$}
\STATE $L_i=\|P_{\theta_p}(h_i,(s_i,a_i))-(s'_i,r_i)\|_2^2$
\ENDFOR
\STATE Prediction loss $\mathcal{L}(\theta_p,\theta_f)=\frac{1}{N}\sum_{i=1}^N L_i$
\STATE Update $P_{\theta_p}$ by minimizing  $\mathcal{L}(\theta_p,\theta_f)$
\STATE Update $F_{\theta_f}$ by minimizing $\mathcal{L}(\theta_p,\theta_f)$ and $I_{vCLUB}$
\end{algorithmic}
\end{algorithm}

\subsection{Architecture of the Hindsight Value Function}
The hindsight value function is designed to leverage the information of the current state and hindsight vector to estimate the expected return. We employed a neural network $v^h_{\theta_v}$ as the hindsight value function. As the input of this function is sequential values with variable length, we naturally consider LSTM\cite{DBLP:journals/neco/HochreiterS97}, a model that operates on sequences. However, inferencing a LSTM for each $v^h(s_t,h^+)$ is computationally expensive. We design the LSTM model to deal with the hindsight vectors in a trajectory at once. As Figure \ref{lstm} shows, the LSTM accumulates information from the hindsight vectors backward along the trajectory. Then the hidden state of LSTM at time $t$ is supposed to be an aggregation of the distilled information of the future after time t. Finally, the hidden state of LSTM $l_t$ and the current state $s_t$ are used to calculate the value estimate. %We give an example algorithm of n-step advantage Actor-Critic with hindsight baseline (with $n=\infty$ for Monte Carl).

\section{Experiment}
%\begin{figure}[htbp]
%\centering
%\includegraphics[width=0.3\linewidth]{env1.pdf}
%\caption{Grid-World}
%\end{figure}

\begin{figure}[t]
\includegraphics[width=\linewidth]{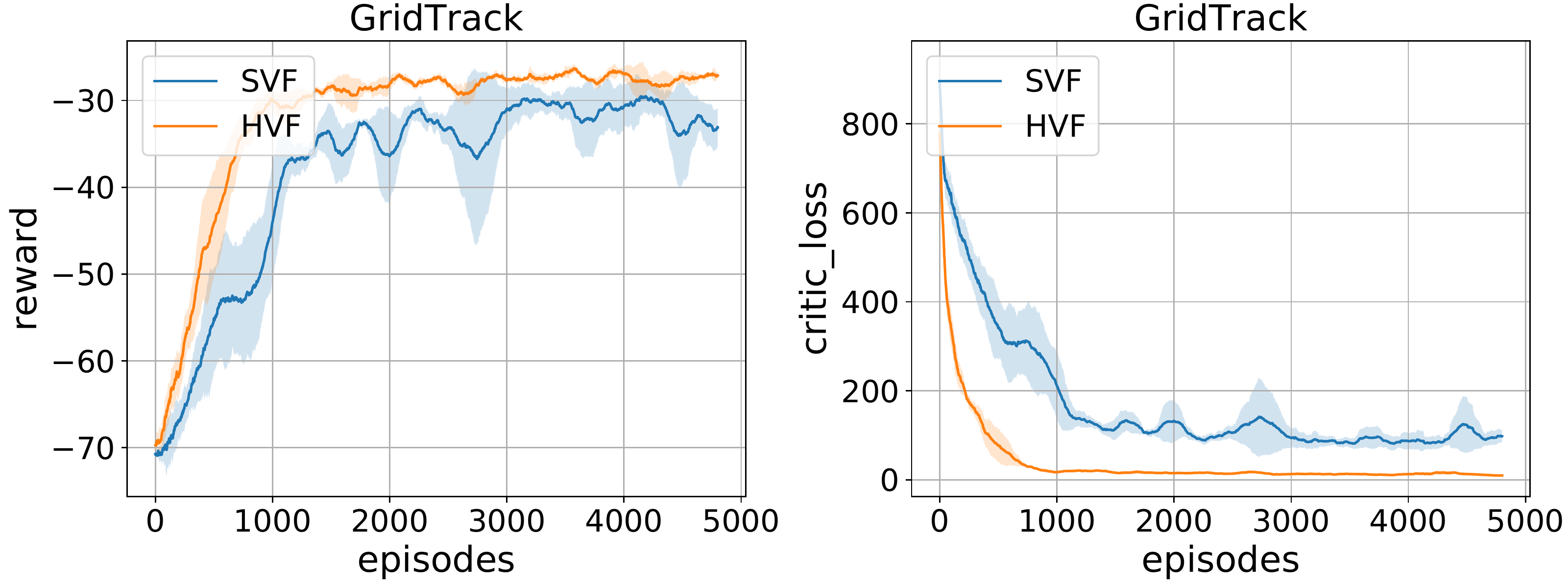}
\caption{Comparison of hindsight value function(HVF) and state value function(SVF) with A2C on GridTrack environment.The shaded area spans one standard deviation.}
\label{gridtrack}
\end{figure}
We conduct experiments on both discrete-action environments and continuous-action environments. To demonstrate the generality and scalability of the hindsight value function, we replace the state value function in A2C\cite{DBLP:conf/icml/MnihBMGLHSK16} and PPO\cite{DBLP:journals/corr/SchulmanWDRK17} with it and show the positive effect. We also carry out experiments to analyze the effect of mutual information constraints.

\begin{figure}[h]
  \centering
  \includegraphics[width=1\linewidth]{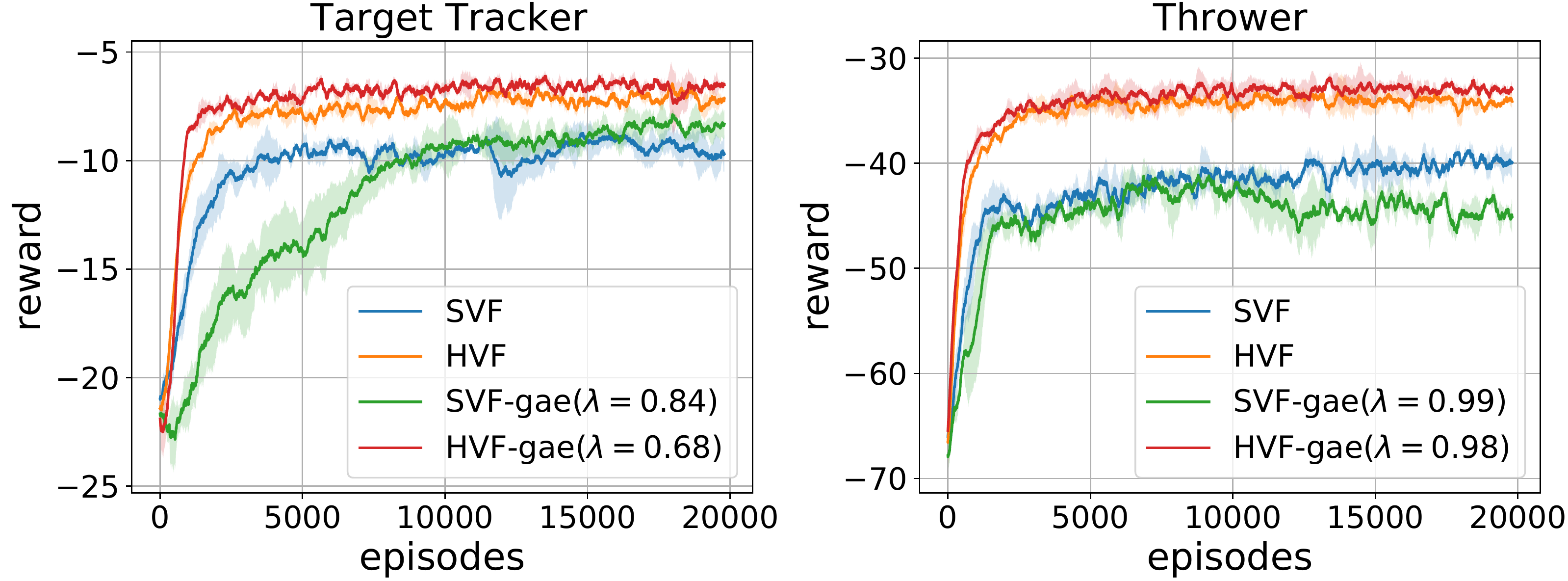}
  \caption{Comparison of hindsight value function(HVF) and state value function(SVF) with PPO on Target Tracker and Thrower environments.}
  \label{reachandthrower}
\end{figure}

\begin{figure}[h]
  \centering
  \includegraphics[width=0.7\linewidth]{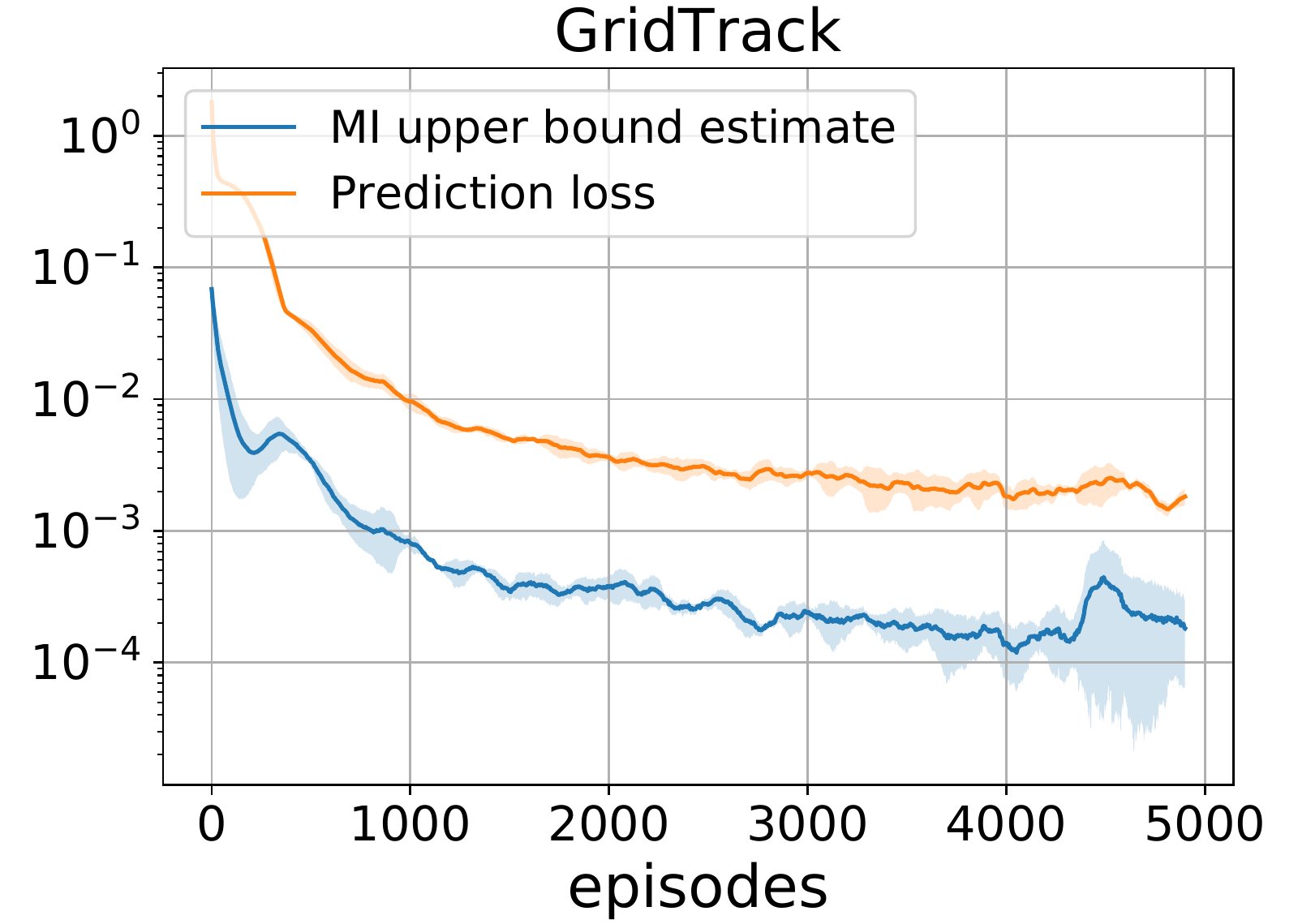}
  \caption{Learning curves of prediction loss and MI upper bound estimate on GridTrack environment.}
  \label{miest}
\end{figure}

\subsection{Environment with Discrete Action Space}
We start from two 8 × 8 versions of the grid-world environment.

\paragraph{Grid track.}
We consider an open grid-world with only bordering walls. There are an agent and a target. 
The agent can move up, down, left, and right, and the target will also move randomly in these four directions. 
We give the agent a dense reward that is the negative squared distance between the agent and the target.

\paragraph{Grid migrate.} 
This environment is similar to Grid Track. The difference is that in this environment, the position of the target is fixed, and the agent will receive a sparse reward when it reaches the target. To make the environment stochastic dynamic, we give the agent a Gaussian random number as a noise reward.   

\paragraph{Settings.}
We set the max length of an episode as 20 for both environments. The discounted factor $\gamma$ is set as 0.99. And we perform the advantage actor-critic(A2C) algorithm on these environments. The actor and critic are implemented as two independent neural networks that are composed of several fully-connected layers. For the hindsight value function, we directly replace the critic with the architecture in Figure \ref{lstm}. Note that for estimating a single value, the new critic maintains the same network architecture but gets an additional input of the LSTM hidden state. Thus, we exclude the influence of the ability of neural networks. We run all these trials with three random seeds. 

\begin{figure*}[htbp]

\centering
\includegraphics[width=0.85\textwidth]{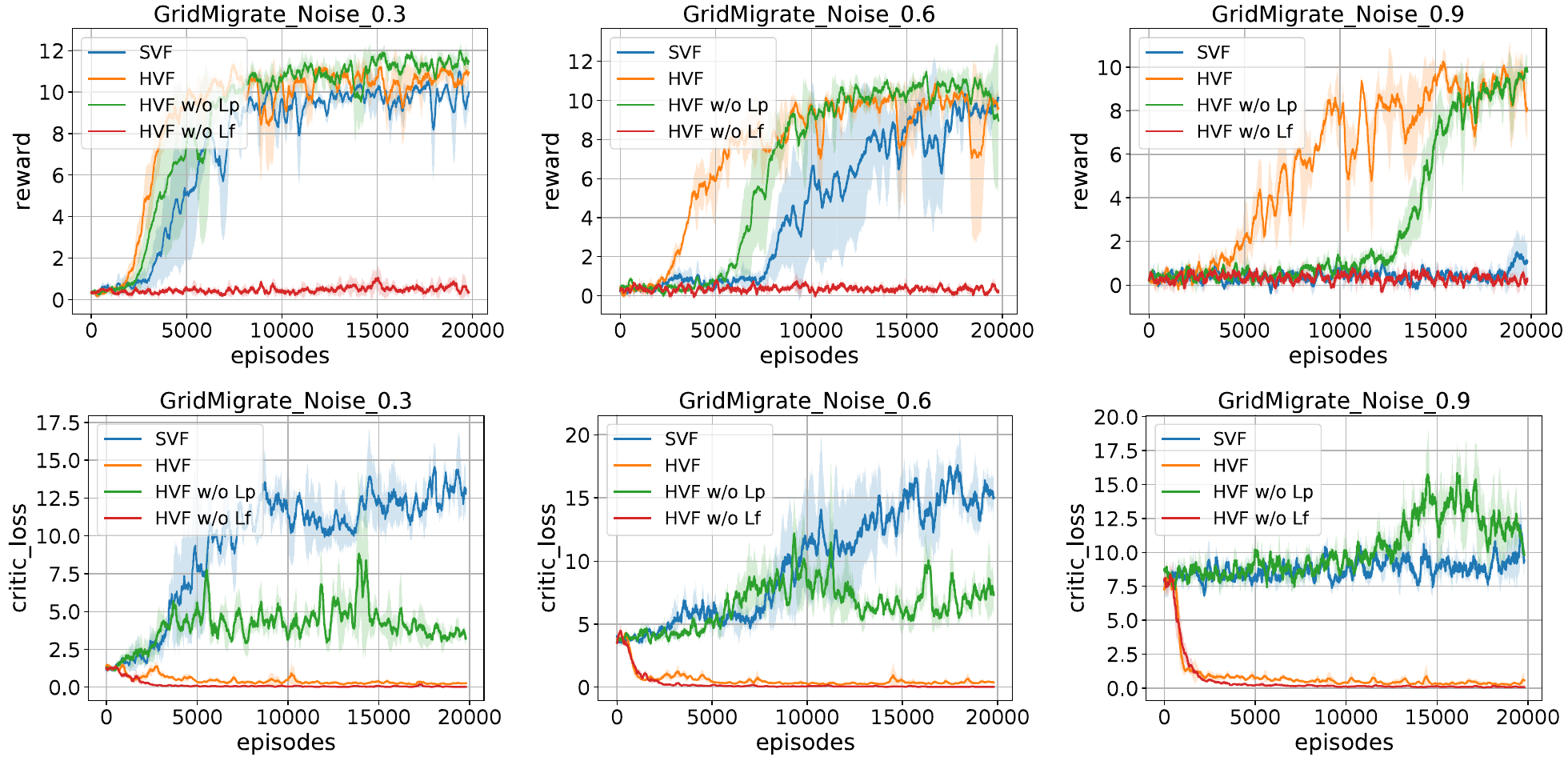}
\caption{Comparison of hindsight value function(HVF) and state value function(SVF) with A2C on Grid Migrate environment under three different noise level.}
\label{migrate}
\end{figure*}

\paragraph{Results.}
Figure \ref{gridtrack} shows the learning curves of the episode reward(the sum of rewards in an episode) and the mean square error of training the critic in the Grid Track environment. Compared to the state value function, the proposed hindsight value function achieves lower critic loss through the whole training process. The hindsight value function gets a better value estimation because it considers the unexpected target moving which will affect the cumulative reward. Thus, the variance of the gradient estimate is reduced, and consequently, the A2C algorithm with hindsight value function as baseline learns a better policy as the episode reward curve shows.
Figure \ref{migrate} shows the learning curves in the Grid Migrate environment. We compare three different noise levels that each correspond to adding Gaussian noise centered at zero
with the labeled standard deviation to the reward at each time step. From left to right, we have the standard deviation as 0.3, 0.6, 0.9. 
As the noise level goes up, the learning of policy becomes slower, and when the standard deviation is 0.9, the standard A2C even can not converge. In contrast, A2C with hindsight value function is robust to noise. The loss of critic training converges to a low level under all noise levels. For all noise levels, A2C with hindsight value function adapts the policy to optima more efficiently compared with standard A2C with state value function.

%\vspace{-0.2in}

\subsection{Environment with Continuous Action Space}
We evaluate the hindsight value function for the continuous-action environment using the MuJoCo robotic simulations in OpenAI Gym\cite{DBLP:journals/corr/BrockmanCPSSTZ16}. We use the FetchReach and Thrower environments. As the state dynamic is deterministic in the MuJoCo simulator, we modify these environments to be stochastic dynamic.  

\paragraph{3-DoF arm target tracker.}
In this environment which is modified from FetchReach, we train a 3-DoF simulated robotics arm to track a randomly moving target. We give the agent a dense reward that is the negative squared distance between the robot hand and the target.

\paragraph{7-DoF arm thrower.}
In this environment which is modified from Thrower, we train a 7-DoF simulated robotics arm to throw a ball to a box on the ground. We give the agent a dense reward that is the negative squared distance between the ball and the box. The box will slightly move randomly when the ball in the robot hand and falling, which makes the environment stochastic dynamic.

\paragraph{Settings.}
We set the max length of an episode as 50 for both environments. The discounted factor $\gamma$ is set as 0.99. And we perform the PPO algorithm on these environments. The network architecture is the same as A2C except for the number of hidden units. We run all these trials with three random seeds. Generalized advantage estimation (GAE) is a method to reduce the variance of gradient estimate at the cost of some bias which is commonly used in PPO implementations. We first turn off GAE by using $\lambda=1$. Then we carry out experiments with GAE turned on. For PPO with GAE enhancement, we try $\lambda = [0.99,0.98,0.96,0.92,0.84,0.68,0.36,0]$ and choose the best one.

\paragraph{Results.}
 Figure \ref{reachandthrower} show the learning curve of episode reward in Target Tracker and Thrower. We find in both environments, PPO with HVF performs better than PPO with SVF. And GAE brings negligible improvement to PPO with SVF. Because when the value function gives a poor value estimate in a stochastic dynamic environment, weighting return estimation with value function bootstrap is meaningless. In the Target Tracker environment, we find PPO-HVF with GAE performs better, which indicates that the proposed hindsight value function may be combined with previous methods of reducing variance to further bring improvement.
 
\subsection{Analysis of Mutual Information}
 In this section, we analyze the effect of constraints of mutual information. In the proposed method, $L_F(\theta_f)$ and $L_P(\theta_f,\theta_p)$ are two additional loss to optimize the mutual information and learn the hindsight value function.
 $L_F(\theta_f)$ is an estimate of the upper bound of mutual information. It is supposed to optimize the gradient estimate towards bias-free. $L_P(\theta_f,\theta_p)$ is a prediction error which maximizes the conditional mutual information $I(h;(s',r)|(s,a))$. It is supposed to optimize the future embedding to contain more information for a better value estimate. Figure \ref{miest} shows the learning curve of $L_P$ and $L_F$ in GridTrack. $L_P$ is optimized to 10e-3, which means the future embeddings contain enough information about the unexpected next state.  As $L_F$ is optimized to 10e-4, the future embeddings can be regarded as independent of previous actions. Then the gradient estimate is unbiased. We verify the effect of $L_F$ and $L_P$ by removing them during training in GridMigrate. As Figure \ref{migrate} shows, when $L_F$ is removed, the policy can not be updated because the expectation of gradient is close to zero as talked at the beginning of Section \ref{hindsight}. When $L_P$ is removed, the policy converges slower but still faster than SVF. Because the embeddings of the future provide limited information, but it is better than no.

\section{Conclusion}
In this paper, we propose a hindsight value function to reduce the variance of gradient estimate in a stochastic dynamic environment. We introduce an information-theoretic approach to obtain an ideally unbiased gradient estimate. The hindsight value function is able to reduce the variance, stabilize the training, and improve the eventual policies in several environments. The approach may inspire future works for reinforcement learning in realistic environments which are mostly stochastic dynamic.

\section*{Acknowledgments}
This work is partially supported by the National Key Research and Development Program of China(under Grant 2018AAA0103300), the NSF of China (under Grants 61925208, 61906179, 62002338, 61732007, 61732002, U19B2019, U20A20227), Beijing Natural Science Foundation (JQ18013), Strategic Priority Research Program of Chinese Academy of Science (XDB32050200, XDC05010300), Beijing Academy of Artificial Intelligence (BAAI) and Beijing Nova Program of  Science and Technology (Z191100001119093) ,Youth Innovation Promotion Association CAS and Xplore Prize.

\bibliographystyle{named}
\bibliography{ijcai21}

\end{document}